\RequirePackage{tikz} 

\documentclass[a4paper, 10pt, conference]{ieeeconf}      

\IEEEoverridecommandlockouts                              

\usepackage{graphicx} 
\usepackage{xspace}
\usepackage{amsmath} 
\usepackage{amssymb}  
\usepackage{bm}
\let\labelindent\relax
\usepackage{enumitem}
\usepackage{hyperref}
\usepackage{nicefrac}

\usepackage[sorting=none]{biblatex}

\def\BibTeX{{\rm B\kern-.05em{\sc i\kern-.025em b}\kern-.08em
    T\kern-.1667em\lower.7ex\hbox{E}\kern-.125emX}}
\title{\LARGE \bf
Preparation of Papers for IEEE Sponsored Conferences \& Symposia*
}

\usetikzlibrary{arrows,calc,positioning,shapes,fit}

\tikzstyle{block} = [rectangle, draw, text centered,
  rounded corners, minimum height=2em]
\tikzstyle{eq} = [circle, draw, minimum height=1.7em, inner sep=1pt] 

\makeatletter
\newcommand{\gettikzxy}[3]{%
  \tikz@scan@one@point\pgfutil@firstofone#1\relax
  \edef#2{\the\pgf@x}%
  \edef#3{\the\pgf@y}%
}
\makeatother

\newcommand{\timeHorizon}{\textrm{\emph{T}}\xspace}
\newcommand{\theGraph}{\textrm{$\mathcal{G}$}}
\newcommand{\nodeSet}{\textrm{$\mathcal{N}$}}
\newcommand{\edgeSet}{\textrm{$\mathcal{E}$}}
\newcommand{\vehicleSet}{\textrm{$\mathcal{V}$}\xspace}
\newcommand{\speed}{\textrm{\emph{$\bar{\Omega}$}}\xspace}
\newcommand{\task}{\textrm{$a$}\xspace}
\newcommand{\taskSet}{\textrm{$\mathcal{A}$}\xspace}

\newcommand{\theSchedule}{\textrm{$\mathcal{S}$}\xspace}
\newcommand{\route}{R}
\newcommand{\Path}{\textrm{$\theta$}}
\newcommand{\EdgeList}{\textrm{$\delta$}}
\newcommand{\Origin}{\textrm{$ \mathcal{D}$}}
\newcommand{\RoutesSet}{\textrm{$\mathcal{R}$}\xspace}

\newcommand{\node}{\textrm{\emph{x}}}
\newcommand{\edge}{\textrm{\emph{y}}}
\newcommand{\edgeLen}[1]{\ensuremath{\lvert #1\rvert}}

\newcommand{\smallPositive}{\textrm{$\mu$}\xspace}

\title{\LARGE \bf
Combining High Level Scheduling and Low Level Control to Manage Fleets of Mobile Robots
}

\author{Sabino Francesco Roselli$^{1}$ and Ze Zhang$^{1}$ and Knut \AA kesson $^{1}$
\thanks{$^{1}$Division of Systems and Control, Department of Electrical Engineering, Chalmers University of Technology, G{\"o}teborg, Sweden{\tt\footnotesize \{rsabino, zhze, knut.akesson\}@chalmers.se}}
\thanks{This work was supported by the Vinnova projects AIHURO (Intelligent m\"anniska-robot-samarbete) and CLOUDS (Intelligent algorithms to support Circular soLutions fOr sUstainable proDuction Systems)}
}

\begin{document}

\maketitle
\thispagestyle{empty}
\pagestyle{empty}

\begin{abstract}
The deployment of mobile robots for material handling in industrial environments requires scalable coordination of large fleets in dynamic settings. This paper presents a two-layer framework that combines high-level scheduling with low-level control. Tasks are assigned and scheduled using the compositional algorithm ComSat, which generates time-parameterized routes for each robot. These schedules are then used by a distributed Model Predictive Control (MPC) system in real time to compute local reference trajectories, accounting for static and dynamic obstacles.
The approach ensures safe, collision-free operation and supports rapid rescheduling in response to disruptions such as robot failures or environmental changes. We evaluate the method in simulated 2D environments with varying road capacities and traffic conditions, demonstrating high task completion rates and robust behavior even under congestion. The modular structure of the framework allows for computational tractability and flexibility, making it suitable for deployment in complex, real-world industrial scenarios.


\end{abstract}

\section{INTRODUCTION}
Mobile robots have become an integral part of material handling in industrial environments, supporting the transport of goods across various areas of a plant \cite{jun2021pickup}. In their simplest form, such systems consist of a small number of robots executing predefined tasks along fixed paths. These setups are relatively easy to manage but offer limited flexibility.
More advanced systems involve larger fleets and allow for more flexibility; robots are able to adjust their trajectory in order to avoid obstacles, both static and dynamic. Moreover, they may change their path altogether if changes in the environment make the current one obsolete or inefficient \cite{fragapane2021planning}.
Such advanced systems offer great advantages and can be deployed in complex scenarios. At the same time, their development poses a few different challenges.

The fleet as a whole needs to be scheduled to execute a list of tasks quickly, or within given time windows, depending on the system; robots need to be assigned to the tasks based on their features that need to match the tasks requirements; the paths computed to travel from and to different locations of the plant also form part of the problem. In fact, it can be assumed that encounters of two or more robots in some areas of the plant will result in deadlocks. Therefore, the choice of paths and the time in which each robot needs to travel them must ensure that such deadlocks do not occur.

Scheduling the fleet is a computationally hard problem, that can only be solved in reasonably short time for small instances limited to a few robots and tasks, as well as a small environment (more details in Section~\ref{sec:experiments}). In such cases, both approximate methods \cite{jun2022scheduling} and exact algorithms \cite{roselli2021solving} can be used effectively. 
For larger systems, problem decomposition may help improve scalability. Specifically, iterating between first assigning robots to tasks and then schedule them along the paths to execute such tasks is a promising approach \cite{roselli2024conflict}. Alternatively, routes and schedules can be computed sequentially for one robot at the time \cite{popolizio2024online}. While these approaches allow for faster computation and can solve larger problem instances, they can't guarantee global optimal solutions.
Another approach to schedule the fleet is to use machine learning algorithms \cite{bogyrbayeva2024machine}; in particular, reinforcement learning has been applied successfully end-to-end \cite{silva2019reinforcement,zhang2020multi} to either drive a single robot, or the entire fleet. 

While scheduling the fleet is necessary to meet the above-mentioned requirements, each robot also needs to be individually controlled to guarantee that it follows its schedule while avoiding collisions with obstacles and other robots. 
Model Predictive Control (MPC) has become a widely used approach for collision-free trajectory tracking, showing effectiveness across various applications \cite{ze_2025_ebmmpc, cbsmpc_2024}. By incorporating obstacles and other robots as constraints and utilizing a receding horizon strategy, MPC can anticipate potential collisions within the planning horizon and generate actions that maintain safe trajectory tracking. By running MPC in a distributed manner \cite{cendismpc_2022}, it can be integrated into other scheduling and coordination methods with good flexibility and scalability.

\begin{figure*}[ht]
    \centering
    \includegraphics[width=.8\linewidth]{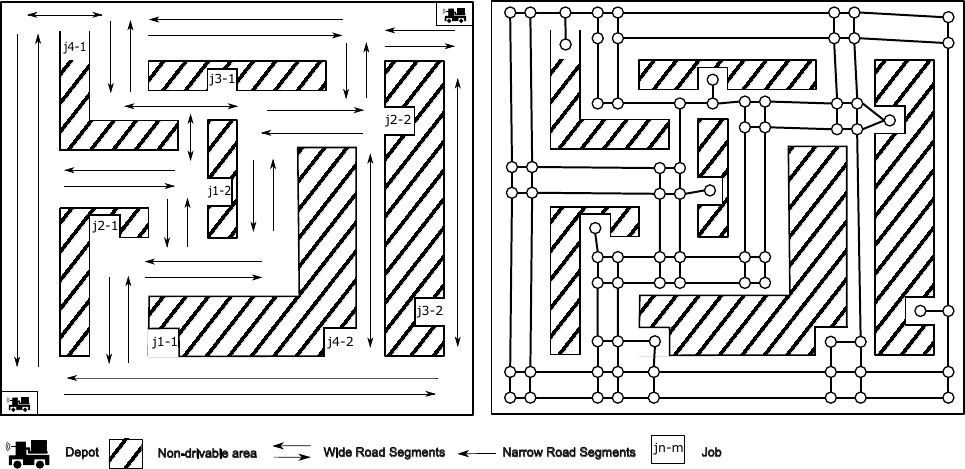}
    \caption{Visualization of an illustrative environment, some roads are wide enough for bidirectional traffic or overtaking in the same direction, which is not possible for other narrow roads. The plant can be abstracted into a graph, as shown in the right sub-figure.}
    \label{fig:scheduler_input}
\end{figure*}

The combination of a high-level centralized fleet scheduler and a low-level distributed controller has proven to be a successful strategy \cite{berndt2021centralized, matos2025efficient, roszkowska2023multi}. In this paper, we present a sequential solution that involves high level scheduling first, i.e., the assignment of tasks to the available robots from the fleet, the design of routes from the robot's initial location, to the locations of the tasks assigned to it, and the scheduling of the robots along their routes, i.e., the relative point in time when the robot is supposed to reach a specific location along its route. This first phase is performed using the compositional algorithm ComSat \cite{roselli2024conflict}. The schedule for each robot becomes the input to the next phase, where a robot-specific local trajectory planner dynamically calculates a local reference trajectory and speed based on the urgency of the current task. Finally, a distributed MPC controller generates a kinematic control action given the real-time reference.

The remainder of the paper is as follows: the problem is formally introduced in Section~\ref{sec:problem_definition}; Section~\ref{sec:fleet_scheduler} introduces the scheduler and Section~\ref{sec:the_bridge} connects it to the model predictive controller, discussed Section~\ref{sec:MPC}; experiments setup and results are reported in Section~\ref{sec:experiments} and, finally, conclusions are drawn in Section~\ref{sec:conclusions}.

\section{PROBLEM AND SYSTEM DEFINITION} \label{sec:problem_definition}

In this section, the fleet management system is defined by dividing it into the scheduler, local planner, and controller. We formalize the problem and provide the interface for each module. In this work, the workspace is assumed to be 2D. We use $t$ for continuous time and $k$ for discrete time, and $t=k\cdot\Delta t$ where $\Delta t$ is the sampling time.

\textbf{Scheduler}:
The schedule is computed over a finite time interval $[0,\timeHorizon]$, where the \emph{Schedule Horizon} \timeHorizon is a fixed and continuous time instant when all tasks should be completed.
As shown on the left side of Fig.~\ref{fig:scheduler_input}, the workspace for mobile robots can be distinguished into drivable (blank areas) and non-drivable (dashed areas). The drivable areas are organized into \emph{road segments} that can be either \emph{narrow} or \emph{wide}; this property of the segments is henceforth named \emph{capacity}. In narrow segments, two robots traveling in the same direction cannot drive next to each other, i.e., overtaking is not possible, and two robots traveling in opposite directions end up in an impasse. In wide segments, the opposite is true. Therefore, as shown on the right side of Fig.~\ref{fig:scheduler_input}, the environment is abstracted into a finite, strongly connected, weighted, directed graph $\theGraph(\nodeSet,\edgeSet)$, where nodes $n\! \in\! \nodeSet$ represent either tasks' locations, depots, or intersections; edges $e\! =\! (n,n')\! \in\! \edgeSet$ represent narrow road segments or lanes of wide road segments; the inverse of an edge $e\! =\! (n,n')$ is denoted $\bar{e}\! =\! (n',n)$, and the length of a road segment is represented by the length of the edge $|e|$. 
   
In a set \vehicleSet of mobile robots, which contains $|\vehicleSet|$ robots, each robot $v \!\in\! \vehicleSet$ moves between locations in the plant at constant speed \speed and performs its tasks. Robots start from their dedicated depot and return either after completing their tasks or intermittently between tasks for battery recharging. As battery-powered vehicles with a limited operating range, their energy consumption is linearly proportional to the distance traveled; they can recharge to full capacity~\cite{schneider2014electric} at their depot, but not at other depots. 
Each task $\task \in \taskSet$, where \taskSet is the set of tasks, represents a place to be visited exactly once by a vehicle to pick up or deliver material. A task $\task$ is always associated with a node where the pickup/delivery takes place, and has a time window $[t^-_{a}, t^+_{a}]$ indicating the earliest and latest time at which the node can be visited. Unless explicitly given, the time window is $[0,\timeHorizon]$. 
Moreover, each task has a precedence list that indicates other tasks that must be executed before the task itself.
This is so since the tasks represent pickups and deliveries, hence a vehicle has to deliver the goods after it has picked them up.

The scheduler processes the graph \theGraph, the task set \taskSet, and the vehicle set \vehicleSet, and produces a schedule \theSchedule, an ordered set of triplets, each specifying the time at which a robot reaches a particular node along its route.

\textbf{Local planner}:
Each robot has its local planner to generate a local reference trajectory at every step. The local planner for robot $v_i$ queries the overall schedule \theSchedule for the specific schedule $s_{i}$ and generates a local reference trajectory $\mathcal{T}^{(i)}_k$ at each time step $k$, consisting of a sequence of $N$ points, where $N$ is the predictive horizon for the MPC controller.

\textbf{Controller}:
For each robot $v_i$, the controller takes the reference trajectory $\mathcal{T}^{(i)}_k$ generated from the corresponding local planner and computes a collision-free action $\bm{u}^{(i)}_k$ accordingly at time step $k$. The controller should avoid collisions with both environmental obstacles and other robots.

\section{FLEET SCHEDULER} \label{sec:fleet_scheduler}

In this section, we introduce the compositional algorithm ComSat \cite{roselli2024conflict} used for the high-level scheduling and coupled with the local planner and controller in the proposed method. To show how the scheduler computes the output \theSchedule based on the input $(\theGraph,\taskSet,\vehicleSet)$, we provide the following definitions:
\begin{itemize}
    \item \textit{Depots}: nodes at which vehicles start and must return to after completing the assigned tasks. 
    \begin{itemize}[label={}]
        \item $\emptyset \subset \Origin \subseteq \nodeSet$: the set of depots.
    \end{itemize}
    \item \emph{Route}: a sequence, starting and ending at the same depot, of unique tasks in-between that may have the same depot embedded: 
    \begin{itemize}[label={}]
        \item $\route_j = \langle d, \task_1, \task_2, \dots, d, \dots, \task_{n-1}, \task_n, d\rangle$ where $d\in \Origin, \task_i,\task_j\in\taskSet$, $\task_i \neq \task_j$ for $i\neq j$, and $n\leq \lvert\taskSet\rvert$ since a route can at most include all tasks.
    \end{itemize}
    \item \emph{Path}: ordered set of unique nodes, used to keep track of how vehicles travel among tasks of routes, since each pair of consecutive tasks is connected by a path.
    \begin{itemize}[label={}]
        \item $\Path_{\task_i,\task_j} = \langle n_1,\ldots,n_m \rangle, \ \task_i,\task_j \in \route, j=i+1, \\ m \leq \lvert \nodeSet \rvert$, $n_i\in\nodeSet,\, i = 1,\ldots,m $
    \end{itemize}
    \item \emph{Edge sequence}: ordered set of unique edges for a given path $\Path_{\task_i,\task_j}$. 
    \begin{itemize}[label={}]
                \item $\EdgeList_{\task_i,\task_j} = \langle e_1,\ldots,e_m \rangle, \ \task_i,\task_j \in \route, j=i+1, \\ m = \vert \Path_{\task_i,\task_j} \vert - 1$, $e_i\in\edgeSet,\, i = 1,\ldots,m $
    \end{itemize}
\end{itemize}

Initially, a set of paths to connect each two tasks/depots in the plant is computed using Dijkstra's algorithm \cite{dijkstra1959note}. Because these are the shortest paths, they can be traveled in the shortest time. Using these paths, a set of routes \RoutesSet is computed by solving the \emph{E-Routing} sub-problem; these routes are designed accounting for the tasks' time windows (if any), their requirements on the type of robot to execute them, as well as the precedence constraints among tasks. Moreover, they take into account the limited operating range of the robots and, if necessary, they involve a visit to the depot to recharge the battery. Because the \emph{E-Routing} sub-problem uses the shortest paths between tasks/depots to compute \RoutesSet, if this problem is infeasible, we know that the overall scheduling problem is infeasible. 
On the other hand, if the \emph{E-Routing} sub-problem is feasible, the set of routes \RoutesSet constitutes the input for the \emph{Capacity Verification}, in which a schedule for each robot to travel along its route is computed; the schedule marks at which time each node in the route should be reached, and accounts for the road segment width, to avoid deadlocks.

In order to adapt the \emph{Capacity Verifier} to the presented framework, one minor adjustment was required. To discuss such an adjustment, we introduce some constraints of the problem, defined using the following variables:
\begin{itemize}[label={},leftmargin=*]
    \item $\node_{\route n}$: non-negative real variable that models the time when a vehicle executing route $\route$ arrives at node $n$;
    \item $\edge_{\route e}$: non-negative real variable that models the time when a vehicle executing route $\route$ starts traversing edge $e$.
\end{itemize}
Also, let $e_n$ be the edge visited before node $n$ and, with some abuse of notation, let $\Path_{R}$ and $\EdgeList_{R}$ be a path and an edge sequence to travel along route $\route$ respectively. Then the relevant constraints for the \emph{Capacity Verification} are:
\begin{flalign}
    &\node_{\route n} \geq \edge_{\route'e^n} + \smallPositive \ \vee \ \node_{\route'n} \geq \edge_{\route e^n} + \smallPositive,  \nonumber\\
    & \qquad \qquad \qquad \quad  \forall \route,\route' \in \RoutesSet, \  \route  \neq \route', \ n \in \Path_{R} \cap \Path_{R} \label{eq:nodes_no_swap} \\
    &\edge_{\route e} \geq \edge_{\route' e} + \smallPositive  \vee   \edge_{\route'e} \geq \edge_{\route e} + \smallPositive, \nonumber \\
    &\qquad \qquad \qquad \quad \forall \route,\route' \in \RoutesSet, \, \route  \neq \route', \ e \in \EdgeList_{R} \cap \EdgeList_{R'} \label{eq:edges_direct}\\ 
    &\edge_{\route e} \geq \edge_{\route'e'} +  \nicefrac{\edgeLen{e'}}{\speed} + \varphi \ \vee \edge_{\route'e'} \geq  \edge_{\route e} + \nicefrac{\edgeLen{e}}{\speed} + \varphi, \nonumber \\
    & \qquad \forall \route,\route' \in \RoutesSet, \ \route \neq \route', \ e \in \EdgeList_{R}, \ e' \in \EdgeList_{R'}, \ e = \bar{e}' \label{eq:edges_inverse}
\end{flalign}

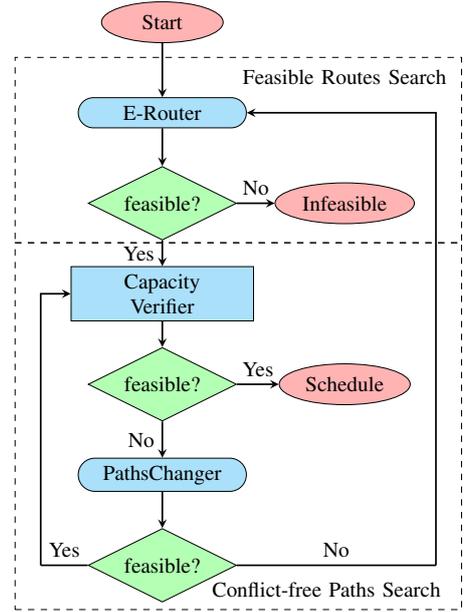
\begin{figure}
    \centering
    \scalebox{0.8}{\tikzstyle{startstop} = [fill=red!30, ellipse, minimum width=2cm, minimum height=0.5cm,text centered, draw=black]

\tikzstyle{io} = [trapezium, trapezium left angle=70, trapezium right angle=110, minimum width=2cm, minimum height=0.5cm, text centered, draw=black]

\tikzstyle{process} = [fill=cyan!30, rectangle, minimum width=3cm, minimum height=0.5cm, text centered, text width=2cm, draw=black]

\tikzstyle{process1} = [fill=cyan!30, rounded rectangle, minimum width=3cm, minimum height=0.5cm, text centered, text width=2cm, draw=black]

\tikzstyle{decision} = [fill=green!30, diamond, aspect=2, text centered, draw=black]

\tikzstyle{arrow} = [thick,->,>=stealth]
\tikzstyle{edge} = [thick]

\begin{tikzpicture}[node distance=1cm]

\node (start) [startstop] {Start};
\node (router) [process1, below of=start, yshift = -0.5cm] {E-Router};
\node (sat1) [decision, below of=router, yshift=-0.5cm] {feasible?};
\node (infeas) [startstop, right of=sat1, xshift=2cm] {Infeasible};
\node (cv) [process, below of=sat1, yshift=-0.5cm] {Capacity Verifier};
\node (sat3) [decision, below of=cv, yshift=-0.5cm] {feasible?};
\node (sch) [startstop, right of=sat3, xshift=2cm] {Schedule};
\node (pc) [process1, below of=sat3, yshift=-0.5cm] {PathsChanger};
\node (sat4) [decision, below of=pc, yshift=-0.5cm] {feasible?};

\coordinate[right of=sat3, xshift=1cm] (coord2);
\coordinate[above of=cv, yshift=-0.27cm] (coord1);
\coordinate[left of=sat4, xshift=-1cm] (coord3);
\coordinate[right of=sat4, xshift=3.5cm] (coord4);

\node[draw,inner xsep=12pt,dashed,fit={(coord1) (coord3) (coord4) (sat4)}]{};

\node[below of=coord4, yshift=0.6cm, xshift=-1.8cm] {Conflict-free Paths Search};

\coordinate[left of=sat1, xshift=-1cm] (coord5);
\coordinate[above of=router, yshift = -0.2cm] (coord6);
\coordinate[right of=infeas, xshift=0.5cm] (coord7);
\coordinate[above of=cv, yshift=-0.04cm] (coord8);

\node[draw,inner xsep=12pt,dashed,fit={(coord5) (coord6) (coord7) (coord8)}]{};

\node[above of=coord7, yshift=1.1cm, xshift=-1.5cm ] {Feasible Routes Search};

\draw [arrow] (start) -- (router);
\draw [arrow] (router) -- (sat1);
\draw [arrow] (sat1) -- node[anchor=east,above] {No} (infeas);
\draw [arrow] (sat1) -- node[anchor=south,left] {Yes} (cv);
\draw [arrow] (cv) -- (sat3);
\draw [arrow] (sat3) -- node[anchor=east, above] {Yes} (sch);
\draw [arrow] (sat3) -- node[anchor=south,left] {No} (pc);
\draw [arrow] (pc) -- (sat4);
\draw [edge] (sat4) -- node[anchor=east,above] {No} (coord4);
\draw [edge] (sat4) -- node[anchor=west,above] {Yes} (coord3);
\draw [arrow] (coord4) |- (router);
\draw [arrow] (coord3) |- (cv);

\end{tikzpicture}}
    \caption{Flowchart of ComSat \cite{roselli2024conflict}.}
    \label{fig:Merged_flowchart}
\end{figure}

Constraints~\eqref{eq:nodes_no_swap}-\eqref{eq:edges_inverse} prevent robots from occupying the same physical location in space at a node or along an edge (see \cite{roselli2024conflict} for the complete model). In the original work, the parameter $\smallPositive$ is a small positive value used to prevent swapping of robots' positions in \eqref{eq:nodes_no_swap} and to forbid robots to traverse the same edge at exactly the same time in \eqref{eq:edges_direct}; $\varphi$ in \eqref{eq:edges_inverse} is equal to zero. However, in this work, $\smallPositive = \varphi$ is a safety margin empirically set to 20 seconds to account for delays in the schedule and prevent robots from occupying the same physical location too close in time. 

If the \emph{Capacity Verification} problem is feasible, a schedule \theSchedule is returned by ComSat; if it is infeasible, new paths (other than the shortest ones computed using Dijkstra's) may be found by solving the \emph{Paths Changing} sub-problem. 

The scheduler iterates between the above-mentioned sub-problems until either a solution \theSchedule is found when the \emph{Capacity Verification} sub-problem is feasible, or the problem is deemed infeasible when the \emph{E-Routing} sub-problem is infeasible. For more details on the workings and performance of this scheduling algorithm, we refer the reader to \cite{roselli2024conflict}. Fig.~\ref{fig:Merged_flowchart} shows how the sub-problems are connected.

\section{BRIDGING SCHEDULER AND CONTROLLER} \label{sec:the_bridge}
The schedule \theSchedule focuses on task allocation and assumes a constant speed of mobile robots to simplify the optimization problem. While in real-world execution, robots may encounter delays due to longer task execution times or traffic obstructions, making it impractical to directly follow the schedule with basic schedule-tracking controllers. To address this, a local planner is employed to bridge the global scheduler with the local control, dynamically adjusting the reference speed and reference trajectory of each robot according to the given schedule.

For each robot $v$ (index $i$ is omitted), its schedule $s$ is a reference path from the start point, passing all intermediate waypoints, to the end point. To achieve real-time performance, a global reference trajectory is precomputed upon receiving the schedule by sampling the reference path under the assumption of maximal velocity $\Omega_{\max}$, rather than generating a local trajectory from scratch at each time step. During operation, an initial reference trajectory $\bar{\mathcal{T}}_k$ is extracted from the global trajectory based on the current location of the robot. Assuming the current time is $t$ and the scheduled arriving time for the next node is $t_n$, the desired speed for the robot at time step $k=t/\Delta t$ will be
$$
\Omega_k = \min\left(\Omega_{\max},\, L_{\text{next}} / \max(0, (t_n-t))\right),
$$
where $L_{\text{next}}$ is the distance to the next node. Then, a new reference trajectory $\mathcal{T}_k$ is generated by resampling the initial one $\bar{\mathcal{T}}_k$ with the ratio $\Omega_k/\Omega_{\max}$.

\section{MODEL PREDICTIVE CONTROL}\label{sec:MPC}
For robot $v_i$, after obtaining the desired speed $\Omega^{(i)}_k$ and reference trajectory $\mathcal{T}^{(i)}_k$, the MPC problem can be solved.
The objective of the MPC problem consists of three parts: reference tracking, obstacle avoidance, and fleet collision avoidance. For reference tracking, at time step $k$,
\begin{equation}
    J_R(k) = ||\!\bm{s}_{k}\!-\tilde{\bm{s}}_{k}\!||^2_{\bm{Q}_s} 
    \!+ ||\!\bm{u}_{k}\!-\!\tilde{\bm{u}}_{k}||^2_{\bm{Q}_u} 
    \!+ ||\!\bm{u}_{k}\!-\!\bm{u}_{k-1}||^2_{\bm{Q}_a},
\end{equation}
where $\bm{s}_{k}$ is the state, $\tilde{\bm{s}}_{k}$ is the reference state from $\mathcal{T}^{(i)}_k$, $\bm{u}_{k}$ is the action, $\tilde{\bm{u}}_{k}$ is the reference action composed by $\Omega^{(i)}_k$, and all $\bm{Q}$ variables are corresponding weight matrices. The obstacle avoidance term $J_{\mathcal{O}}(k)$ is based on the indication of the robot being inside or outside of a polygonal obstacle \cite{cendismpc_2022}. The fleet collision avoidance is defined as
\begin{equation}
    J_F(j) = \max\left[0, Q_f\cdot\left(d_{\text{fleet}}-||\bm{z}_{k}^{(i)}-\bm{z}_{k}^{(j)}|| \right)^2\right],
\end{equation}
where $\bm{z}^{(i)}$ is the position of robot $v_i$, $d_{\text{fleet}}$ is the safe distance between robots, and $Q_f$ is the weight parameter.

Assuming the current time step $k=0$, the complete MPC problem can be formulated as
\begin{align}
    \min_{\bm{u}_{0:N-1}} & \sum_{{k}=0}^{N-1} \bigg[ J_R(k) + J_{\mathcal{O}}(k) + \sum_{j=1,j\ne i}^{|\mathcal{V}|}J_F(j) \bigg],  \\
    \text{s.t.} \quad 
    & \bm{s}_{k+1} = f(\bm{s}_{k}, \bm{u}_{k}), \, \forall k = 0, \ldots, N-1, \\
    & \bm{u}_{k} \in [\bm{u}_{\text{min}}, \bm{u}_{\text{max}}], \, \forall k = 0, \ldots, N-1, \\
    & \Dot{\bm{u}}_{k} \in [\Dot{\bm{u}}_{\text{min}}, \Dot{\bm{u}}_{\text{max}}], \, \forall k = 0, \ldots, N-1, \\
    & \bm{z}_{k} \notin \mathcal{O}, \, \forall k = 0, \ldots, N-1,
\label{eq:mpc}
\end{align}
where $f(\cdot)$ is the motion model, $\Dot{\bm{u}}_{k}$ is the derivative of the action, and $(\bm{u}_{\text{min}/\text{max}}, \Dot{\bm{u}}_{\text{min}/\text{max}})$ are physical limits of the robot's action.

\begin{figure}
    \centering
    \includegraphics[width=0.4\textwidth]{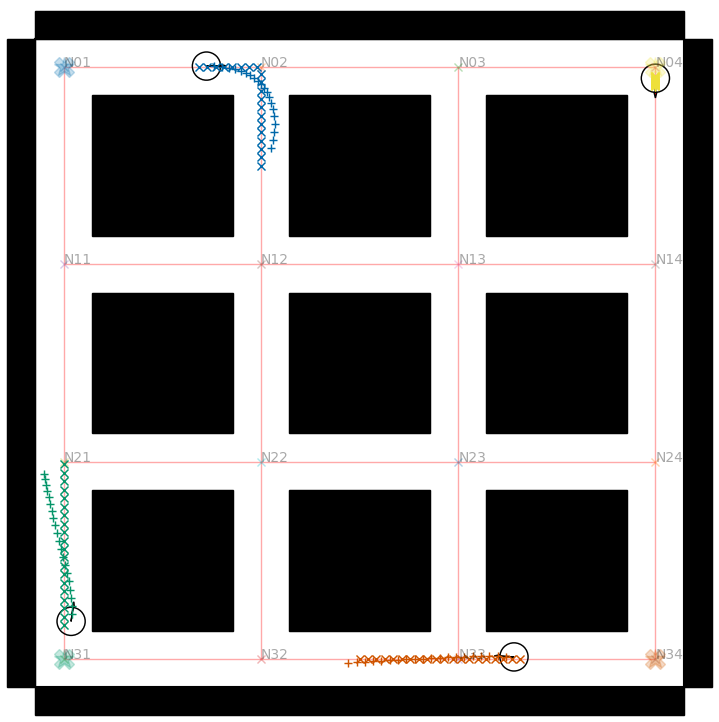}
    \caption{Map of the small environment with four horizontal and four vertical narrow road segments. The map is abstracted into a graph (nodes from \emph{N01} to \emph{N34} and orange edges), and the robots are represented by circles, with their reference trajectories marked by crosses ($\times$) of different colors and predicted trajectories marked by pluses ($+$).}
    \label{fig:Small4Robots}
\end{figure}

\section{EXPERIMENTS}\label{sec:experiments}

In this section we present the experimental setup developed to test the proposed framework, as well as the results achieved by it. The \emph{small} setup is a grid-like environment with four horizontal and four vertical narrow road segments, abstracted into a graph with 16 nodes, as shown in Fig.~\ref{fig:Small4Robots}. In this setup, four robots are initially located at the four corners of the map (the depots), and each of them has to travel to three other locations before returning to its depot. 
For instance, robot \emph{A4}, represented in yellow and initially located in the top-right corner, corresponding to Node \emph{N04} has to travel to the locations \emph{N23}, \emph{N21}, and \emph{N24}, in this order, before going back to \emph{N04}. 

\begin{figure}[ht]
    \centering
    \includegraphics[width=0.48\textwidth]{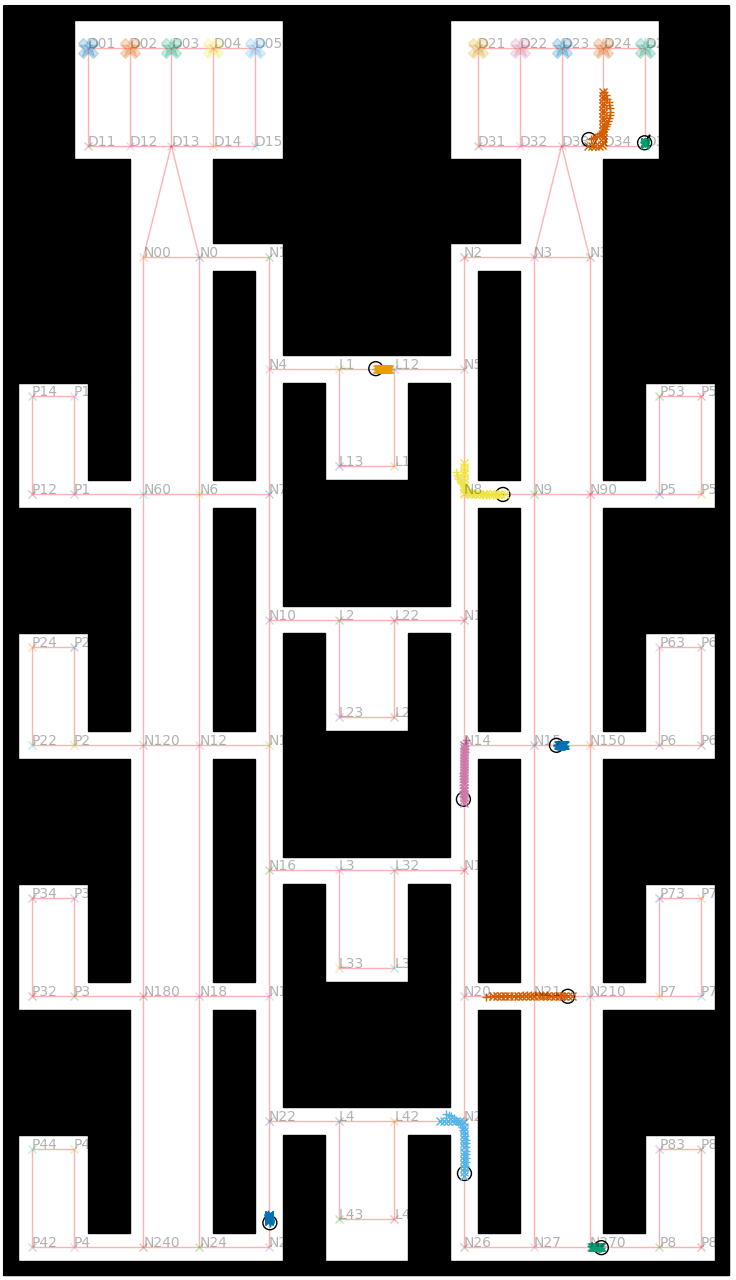}
    \caption{Map of the large environment mimicking a manufacturing/assembly plant layout. The main line where goods need to be shipped is in the middle, with four workstations, while four kitting areas on each side are where material is picked up. The robots' depots are located in two areas at the top. The map is abstracted into a graph, and the robots are represented by the circles, with their planned trajectories marked by \emph{x} of different colors.}
    \label{fig:Large10Robots}
\end{figure}

A second experiment, shown in Fig.~\ref{fig:Large10Robots} (last page) is set up using a map that mimics a factory layout, in which an assembly line is located in the middle and on each side of it there is a set of kitting areas where the robots, dispatched from two depot areas, pick up the material to deliver to the assembly line. All road segments are narrow, with the exception of the two main ones running parallel to the assembly line. The map is abstracted into a graph with 106 nodes and 292 edges. In this set up, ten robots are initially located in the two depot areas, five in each; each robot starts and returns to its own depot, after visiting one of the kitting areas (pickup) and one of the work stations in the assembly line (delivery).

\begin{figure}
    \centering
    \includegraphics[width=0.48\textwidth]{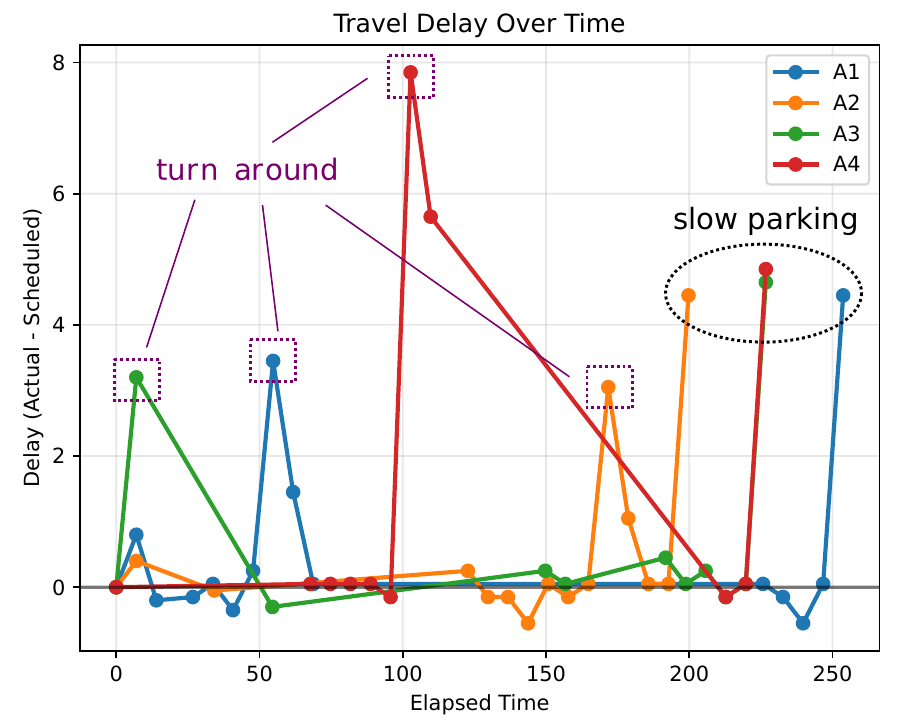}
    \vspace{-5mm}
    \caption{Execution delays (in Seconds) of the four robots in the small environment. The delay, positive when the robot is late or negative otherwise, is the difference between the actual arrival time at a node versus the scheduled arrival time.}
    \label{fig:delayAnalysisSmall4}
\end{figure}

The scheduler uses the state-of-the-art MILP solver Gurobi 12.0.1 \cite{gurobi} to solve the \emph{E-Routing} and \emph{Path Changing} sub-problems and the state-of-the-art SMT solver Z3 4.14.1 \cite{de2008z3} to solve the \emph{Capacity Verification} sub-problem. 
The MPC problem is solved via the PANOC algorithm \cite{Stella_2017_panoc}, which is a real-time nonlinear optimal control solver.
Experiments are run on a MacBook Pro with 18GB of RAM and M3 Pro core running Sequoia 15.5.\footnote{The implementation of the proposed framework, including the scheduler based on ComSat, the MPC, and the problem instances, is available at \url{https://github.com/Woodenonez/TrajPlan-ScheMPC}}

Both experiments show that the robots are able to accomplish their tasks and return to their depots in due time. In Fig.~\ref{fig:delayAnalysisSmall4} is reported the delay of each robot to reach each node of its route for the experiment in the small environment; this delay is computed by logging when the robot actually reaches a certain node during the simulation and comparing it to the scheduled time in which the robot should reach such node. 
The delay for all robots ranges from 0 to 5 seconds on average, with a peak of almost 8 seconds for robot \emph{A4}. Delays happen when robots need to perform 180 degrees changes of directions, which are non instantaneous; however, the controller increases the speed of the robot when falling behind schedule on a road segment and is able to compensate. The delay all robots experience when reaching the last node of their route is instead due to an intended slow parking behaviour. 

\begin{figure}
    \centering
    \includegraphics[width=0.49\textwidth]{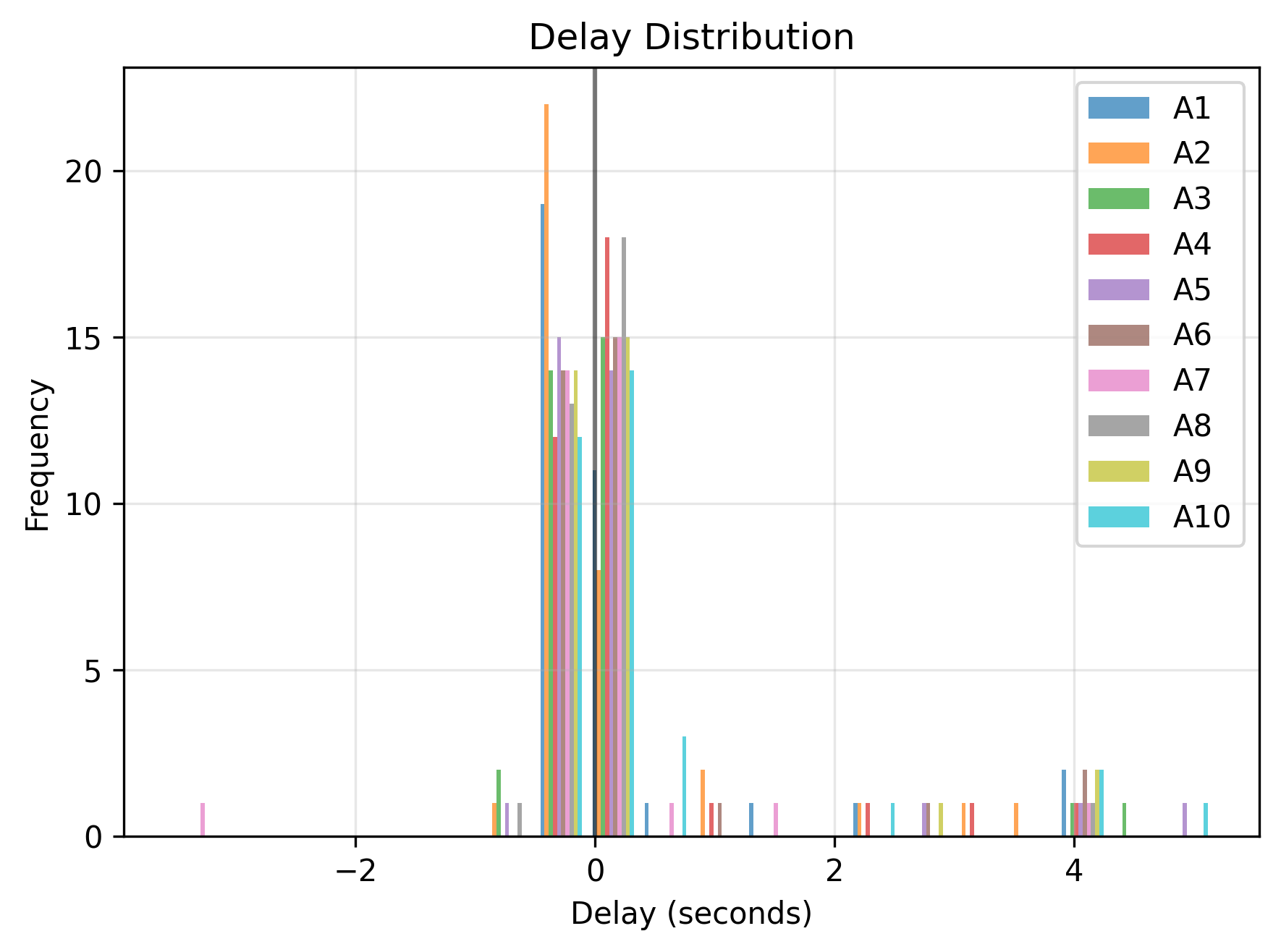}
    \vspace{-5mm}
    \caption{Delay distribution for ten robots in the large environment.}
    \label{fig:delayAnalysisLarge10}
\end{figure}

A similar result is achieved in the large environment with ten robots. Fig.~\ref{fig:delayAnalysisLarge10} shows the delay distribution for each robot. The vast majority of arrival times at the node fall within 1 second before or after the expected arrival time, with a maximum lateness below five seconds. This consistency across environments of varying scale suggests that the control strategy scales effectively, maintaining tight synchronization with scheduled plans even as the number of agents and route complexity increase. 

We compared the presented method with two alternatives; one, in which the MPC on each robot is replaced with a simple \emph{reference tracker} that does not account for the other robots' presence, but only for static obstacles. In this case, collisions happened if two robots were scheduled to travel through an intersection at similar time. The other alternative used the MPC presented in Section~\ref{sec:MPC} and the routes computed by the scheduler, but no time-schedule, i.e., the robots are traveling along the routes as fast as they can. In this case, the lack of centralized scheduling resulted in collisions for every single simulation. 

\section{CONCLUSIONS AND FUTURE WORK}\label{sec:conclusions}

This work presents an integrated framework that bridges high-level scheduling and low-level control to manage fleets of mobile robots operating in dynamic and constrained environments. By combining the compositional scheduling algorithm ComSat with a distributed Model Predictive Control (MPC), we demonstrate a scalable and responsive system capable of handling both long-term planning and real-time execution challenges. The proposed method first computes conflict-free schedules, then adapts these at runtime through a local planner and MPC-based controller, ensuring safe and timely task completion despite disturbances such as delays or congestion.

The experimental results validate the practicality of our approach. In both small-scale and large-scale scenarios, the robots successfully complete their assigned tasks and return to their depots with minimal deviation from the schedule. Importantly, the use of MPC enables robots to recover from potential delays by dynamically adjusting their trajectories, highlighting the method’s robustness and flexibility.

Moreover, the framework is modular, allowing for future extension with additional constraints, such as energy optimization, learning-based adaptations, or human-robot interaction models. As mobile robot fleets become increasingly common in industrial and logistics settings, this work provides a solid foundation for building efficient, adaptable, and scalable robotic management systems

The obvious next step is to implement an experimental setup with real robots to validate the proposed method under realistic conditions. This will allow evaluation of its robustness against uncertainties such as sensor noise, communication delays, and environmental interferences, providing insight into its practical feasibility beyond simulation.

\printbibliography

\end{document}